\documentclass[letterpaper, 10 pt, conference]{ieeeconf}  

\IEEEoverridecommandlockouts                              

\usepackage{amsmath} 
\usepackage{amssymb}  
\usepackage{xcolor}
\usepackage{algorithm}
\usepackage{algpseudocode}
\usepackage{balance}
\usepackage{wrapfig}
\usepackage{array,multirow,graphicx}
\usepackage{subcaption}
\usepackage{booktabs}
\usepackage{verbatim}
\usepackage{lipsum}
\usepackage{mathtools}
\usepackage{microtype}            
\usepackage{url} 
\usepackage{etoolbox}
\apptocmd{\thebibliography}{\sloppy}{}{} 
\algrenewcommand\algorithmicrequire{\textbf{Inputs:}}
\algrenewcommand\algorithmicensure{\textbf{Output:}}

\newif\ifnotes
\notesfalse 

\ifnotes
  \newcommand{\rmm}[1]{\textcolor{blue}{RMM:~#1}}
  \newcommand{\sateesh}[1]{\textcolor{cyan}{Sateesh: #1}}
  \newcommand{\taijing}[1]{\textcolor{violet}{Taijing: #1}}
  \newcommand{\jhx}[1]{\textcolor{purple}{Junhong: #1}}
  \newcommand{\jb}[1]{\textcolor{red}{Joydeep:~#1}}
\else
  \newcommand{\rmm}[1]{}
  \newcommand{\sateesh}[1]{}
  \newcommand{\taijing}[1]{}
  \newcommand{\jhx}[1]{}
  \newcommand{\jb}[1]{}
\fi
\newcommand{\methodname}{STAR}
\newcommand{\methodnamelong}{SpatioTemporal Active Retrieval}
\newcommand{\benchmarkname}{STARBench}

\usepackage{cite} 
\usepackage{placeins}

\title{\LARGE \bf
Searching in Space and Time: Unified Memory-Action Loops for Open-World Object Retrieval
}

\begin{document}

\author{
    Taijing Chen$^{1}$, 
    Sateesh Kumar$^{1}$, 
    Junhong Xu$^{1}$, 
    Georgios Pavlakos$^{1}$, 
    Joydeep Biswas$^{1,*}$, 
    Roberto Martín-Martín$^{1,*}$%
    \thanks{$^{1}$The authors are with the Department of Computer Science, The University of Texas at Austin, Austin, TX, USA. \texttt{\{taijingchen, sateesh, jh.xu, pavlakos, joydeepb, robertomm\}@utexas.edu}}
    \thanks{S. Kumar is funded by the Amazon AI PhD fellowship.}
    \thanks{This work was supported by the Amazon Research Awards (ARA) program.  Any opinions, findings, and conclusions expressed in this material are those of the authors and do not necessarily reflect the views of the sponsors.}
}

  \maketitle
\thispagestyle{empty}
\pagestyle{empty}

\begin{abstract}

Service robots must retrieve objects in dynamic, open-world settings where requests may reference attributes (“the red mug”), spatial context (“the mug on the table”), or past states (“the mug that was here yesterday”).
Existing approaches capture only parts of this problem: scene graphs capture spatial relations but ignore temporal grounding, temporal reasoning methods model dynamics but do not support embodied interaction, and dynamic scene graphs handle both but remain closed-world with fixed vocabularies.
We present \methodname{} (\textit{\methodnamelong{}}), a framework that unifies memory queries and embodied actions within a single decision loop. STAR leverages non-parametric long-term memory and a working memory to support efficient recall, and uses a vision-language model to select either temporal or spatial actions at each step. We introduce STARBench, a benchmark of spatiotemporal object search tasks across simulated and real environments. Experiments in STARBench and on a Tiago robot show that STAR consistently outperforms scene-graph and memory-only baselines, demonstrating the benefits of treating search in time and search in space as a unified problem. For more information: https://amrl.cs.utexas.edu/STAR.

\end{abstract}





\section{INTRODUCTION}

We are interested in the problem of \emph{open-world object retrieval}, where a service mobile robot is asked to retrieve arbitrary objects referred to through combinations of open-vocabulary appearance (``the red mug''), spatial properties (``the mug on the dining table''), and temporal properties (``the mug that was on the coffee table yesterday''). This problem comes with several challenges. First, since the robot does not know in advance which objects it will be asked to retrieve, it cannot explicitly track and remember all possible solutions. Second, even if it has observed the relevant object before, that object may no longer be at its most recently observed location. Third, retrieving an object may require the robot to interact with the environment, such as opening drawers or cabinet doors.

\begin{figure}
    \centering
    \includegraphics[width=1.0\linewidth]{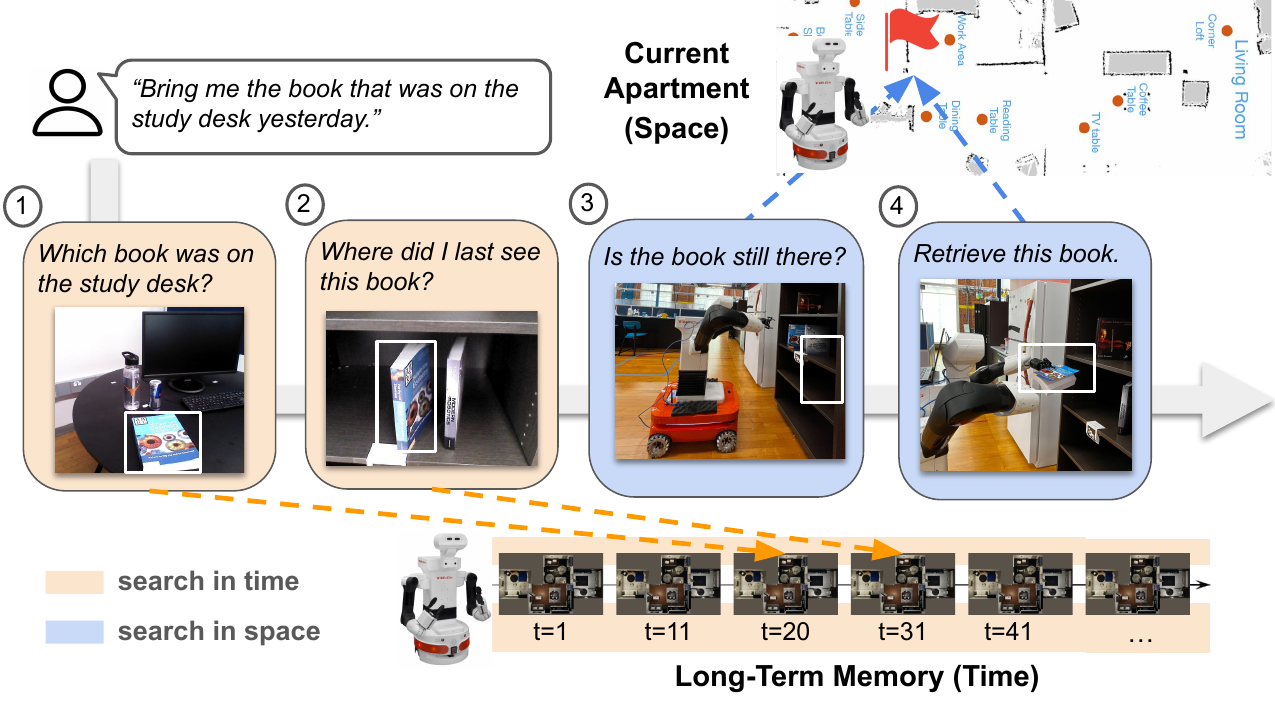}
     \caption{A fundamental skill for a robot is to retrieve desired objects, often specified through a combination of spatial and temporal references (e.g., \textit{``bring the book that was on the desk yesterday''}). \methodname{} is a framework that integrates long-term memory with an agentic search engine, enabling the robot to decide whether to search in time (recall past observations) or in space (probe the current environment), producing action sequences (2 to 6) to obtain the necessary information to retrieve the correct object.}
    \label{fig:pf}
\end{figure}

We note that solving this problem requires reasoning jointly about \emph{space} (where the object is, how to reach it, what to manipulate to access it) and \emph{time} (when it was last observed, how often it appears there, and whether it may have moved).
Crucially, the robot must do so over an open set of objects, natural language referring expressions, and arbitrary temporal dynamics. 
Existing approaches to solve this problem focus either on searching in space or in time, but rarely both. Approaches to searching in space include Object Maps~\cite{johnson2015image}, Scene Graphs~\cite{armeni20193d}, and most recently Open-Vocabulary Scene Graphs~\cite{ConceptGraph2024}. Approaches to searching in time include video retrieval and grounded temporal reasoning methods~\cite{arora2024g2trgeneralizedgroundedtemporal}. A few recent approaches explore both space and time, such as dynamic scene graphs~\cite{DovSG2025,pham2025tesgnntemporalequivariantscene}, which however assume a closed set of objects and fixed relational vocabularies. To date, no prior work enables a robot to retrieve arbitrary, open-world objects described in natural language with both spatial and temporal references.

In this paper, we introduce STAR (SpatioTemporal Active Retrieval), a unified approach that integrates spatial and temporal search for open-world object retrieval. 
Our key insight is to treat memory queries (searching in \emph{time}) and physical actions (searching in \emph{space}) as elements of a unified action space inside an active interaction decision-making loop. 
This design combines the strengths of prior approaches: as in retrieval-augmented agents, it can leverage memory of past observations; and as in reactive agents, it can act in the current environment to gather new perceptual evidence and execute navigation or manipulation actions. 
To reason about past observations, STAR maintains a \emph{long-term memory} of its observations in the world represented as a lightweight sparse language-captioned set of observations inspired by ReMEmbR~\cite{anwar2025remembr}, coupled with periodic full visual snapshots of the robot's observation. When given a new object retrieval task, STAR uses a large language model (LLM) in a decision-making loop to decide which spatial or temporal actions to execute, compiling results in a \emph{working memory} that is used to decide on successive actions until it ascertains that it has retrieved the object that the user requested.

Evaluating such a capability requires benchmarks that capture long-term, dynamic environments. Since existing benchmarks for object search either assume static scenes or focus on passive reasoning over recorded videos for Visual Q\&A, we developed STARBench, a benchmark for spatiotemporal object search in dynamic households. STARBench spans 360 tasks across visible, interactive, and commonsense settings and five instruction families. STAR consistently outperforms baselines across all task types and transfers to a Tiago robot in a mock apartment, 
with pronounced advantages on tasks requiring reasoning about object properties and references to past observations.

Our contributions are threefold:
\begin{itemize}
    \item We propose STAR, a framework that unifies search in \emph{time} (memory retrieval) and search in \emph{space} (embodied actions) for open-world object retrieval
    \item  We introduce STARBench, a benchmark for spatiotemporal object search in dynamic household environments
    \item We show STAR outperforms scene-graph and temporal-search-only baselines across all task types in STARBench, and demonstrates successful transfer to a Tiago robot in a mock apartment
\end{itemize}
\section{Related Work}

\methodname{} tackles embodied object search by unifying interactions with dynamic memory in a framework that searches in both space and time. We discuss connections to prior work in object search, benchmarking, LLM-based agents, and memory in robotics.

\noindent
\textbf{Object Search.} Methods for object search aim to find a desired object through navigation and/or manipulation. 
In \emph{object navigation}, robots move through the environment to detect a visible target object. 
While early works relied on geometric reasoning~\cite{kuipers1991robot}, more recent approaches use semantic representations such as scene graphs~\cite{kumar2021gcexp, amiri2022reasoning,santos2022deep} or learned priors~\cite{chang2020semantic, liang2021sscnav, wortsman2019learning, yang2018visual}. 
To deal with the continuously changing nature of human-populated environments, recent approaches integrate a dynamically updating representation of predefined or open-vocabulary objects~\cite{DovSG2025,chang2023goat, pham2025tesgnntemporalequivariantscene, lingelbach2023task, kurenkov2023modeling}.
However, such methods assume the critical information is only in the current state; they cannot deal with retrospective queries such as ``bring me the book that was here yesterday.'' 
Analogously, in \textit{interactive object search} (i.e., Mechanical Search~\cite{danielczuk2019mechanical}) robots exploit manipulation actions ---opening cabinets or drawers, or removing occluding objects--- to find the target object~\cite{honerkamp2024language, schmalstieg2023learning,huang2022visual, danielczuk2019mechanical, kurenkov2021semantic, kurenkov2020visuomotor, huang2021mechanical, huang2022mechanical, huang2022mechanica2}, making decisions with either a direct sensorimotor policy or a scene representation restricted to a predefined vocabulary of objects and updated to reflect only the current state.
While \methodname{} also uses interactions to search, it leverages an open-vocabulary memory representation with history that enables it to combine interactions (i.e., interactive searches in space) with searches in time.

\noindent
\textbf{Benchmarking Object Search}. Early benchmarks to evaluate object search methods originated in the computer vision community under the umbrella of EmbodiedQA~\cite{das2018embodied} and VisualQA~\cite{antol2015vqa,gordon2018iqa}.
They evaluate embodied agents on broad tasks that combine navigation, perception, and language understanding, with object search included as one of the components. 
Instruction-following suites like ALFRED~\cite{shridhar2020alfred} extend this to household tasks requiring both navigation and manipulation. 
More recent work targets object navigation more directly on a predefined set of objects~\cite{ramrakhya2022habitat} or extended to open-vocabulary settings~\cite{gadre2023cows,yokoyama2024hm3d}. 
While these benchmarks advance object search, they only evaluate search tasks based on the current state of a static environment; no existing benchmark includes tasks that require the agent to reason about previous states of a dynamically changing world to retrieve the correct item, a common situation in assistive robotics. 
\emph{\benchmarkname{}} fills this gap with a benchmark for spatiotemporal interactive object search in dynamic environments.

\noindent
\textbf{LLM Agents for Robotics.} The impressive advances in the last years in large language and vision–language models (LLM/VLM) have enabled multiple applications of their common-sense reasoning as planners and controllers in robotics~\cite{ahn2022can,liang2022code,shah2023lm,huang2022inner, rana2023sayplan,song2023llm,singh2022progprompt, shah2025bumble}. 
As pioneers, \textit{Code as Policies}~\cite{liang2022code} generates executable robot programs directly from natural language, leveraging the structure of code to integrate perception and control primitives while supporting iterative refinement through execution feedback, while LM-Nav~\cite{shah2023lm} combines large-scale language grounding with pretrained visual recognition and navigation modules, parsing instructions into semantic waypoints and re-planning as needed based on perceptual feedback. These systems illustrate how LLM/VLM can enable plan–execute–check loops by coupling reasoning with perception and control, yet they largely operate over the current scene and short contexts. In contrast, \methodname{} leverages an LLM to generate code as memory queries and embodied actions within the same decision loop, enabling agents to reason jointly over past and present conditions.

\noindent
\textbf{Memory and Scene Representations for Robot Navigation.}
Robotics often relies on memory to represent objects, places, and their relations. 
Parametric structures such as scene graphs and their open-vocabulary or dynamic extensions~\cite{ConceptGraph2024, NLMap2023, koch2024open3dsg} capture the current world state but overwrite past information. 
A parallel line of work in long-term semantic mapping develops persistent object-level maps that are updated across sessions to improve robustness in navigation~\cite{SalasMoreno2013SLAMpp,McCormac2017SemanticFusion,McCormac2018FusionPP,Narita2019PanopticFusion,Rosinol2020Kimera,Martins2024OVOSLAM,Adkins2023ObViSLAM}, yet these maps also emphasize the current belief rather than retaining accessible histories of past states. 
Due to the complexity of building and maintaining these explicit representations, recent non-parametric approaches store instead raw observations for retrieval, as in ReMEmbR~\cite{anwar2025remembr} or Embodied-RAG~\cite{xie2025embodiedraggeneralnonparametricembodied}. 
While these systems demonstrate the value of retrieval, memory typically remains an external module consulted before action. 
Inspired by the flexibility to handle open-vocabulary queries, \methodname{} integrates a non-parametric memory into the action space itself, enabling agents to recall earlier states, use them even when they differ from the present, and adapt behavior when discrepancies arise.

\section{Problem Statement}
In this section, we describe the problem of open-world object retrieval in dynamic, partially observable environments.
Let $t \in \{0, 1, 2, \ldots\}$ denote the discrete timestep and 
$e_t \in \mathcal{E}$ the environment state at $t$, e.g., a configuration of an apartment, which may change over time due to exogenous factors, e.g., human activities.
At each step, the available information $s_t = (t, o_t, x_t)$ to the robot consists of the current timestep $t$ and a sensory observation $o_t \in \mathcal{O}$ of $e_t$ at the robot pose $x_t$.

At task time $t=T$, the robot receives an open-ended natural-language instruction $\ell \in \mathcal{L}$ from the user, specifying a target object to be retrieved. 
The instruction may describe the target via attributes, spatial relations to other objects, or temporal cues.
As a result, the space of instructions $\mathcal{L}$ is vast and diverse.
Some instructions contain pointed temporal reference (``yesterday"), others have habitual patterns (``usually on the desk"), and some depend on objects' attributes and spatial context (``the blue mug in the kitchen). 
We assume that before the task time, the robot patrols the environment and receives observations $S_{\leq T} = \{s_i\}_{i=0}^T$.
To complete the task, the robot needs to maintain $S_{\leq T}$ to understand which object $\ell$ refers to.
Since observations are high-dimensional and histories may span days, the robot must \emph{store and index} past observations into queryable representations for efficient retrieval at task time.
We formalize this process with a memory construction operation: $M = \Phi(S_{\leq T})$ where $\Phi$ maps the history of observations into a long-term memory $M$.
Different works realize this construction differently, such as structured scene graphs~\cite{armeni20193d} or semantic mapping~\cite{SalasMoreno2013SLAMpp, McCormac2017SemanticFusion, McCormac2018FusionPP, Narita2019PanopticFusion, Rosinol2020Kimera, Martins2024OVOSLAM, Adkins2023ObViSLAM}.
Importantly, since the variations of the user instructions are enormous and are unknown before the task time, memory must be constructed in a task-agnostic way.

After receiving $\ell$ at time $T$, the robot executes for at most $K$ steps to retrieve the requested object.
Let $S_{T:T+k} = \{s_i\}_{i=T}^{T+k}$ denote the observation stream accumulated since~$T$, where $k \in \{1, \ldots, K\}$ denotes the timestep of the policy execution.
For simplicity, we will use timestep $k$ to represent timestep $T+k$ whenever the context is clear.
At each step, the policy selects an action based on the instruction, observation stream, and the long-term memory: 
\begin{align}
\label{eqn:full-problem}
a_k \sim \pi(a \mid l, M, S_{T:T+k}),
\end{align}
which moves the robot to a new pose and obtains a new observation.
The robot's objective is to reach the instructed object before the {K}-step budget expires, preferably in fewer steps.
Because the environment may have changed, i.e., $e_{T+k}$ may differ from $(e_1, ..., e_T)$, long-term memory $M$ may be stale.
The robot needs to decide how much to rely on information stored in $M$ versus how much new evidence to acquire: selectively recall from $M$ to form hypotheses about likely object locations, and, when those expectations are not met, act to gather new observations to refine where to search next, so the robot reaches the goal within the K-step budget.
\section{\methodname{}: \methodnamelong{}}

\begin{figure*}[t]
    \centering
    \includegraphics[width=1.0\linewidth]{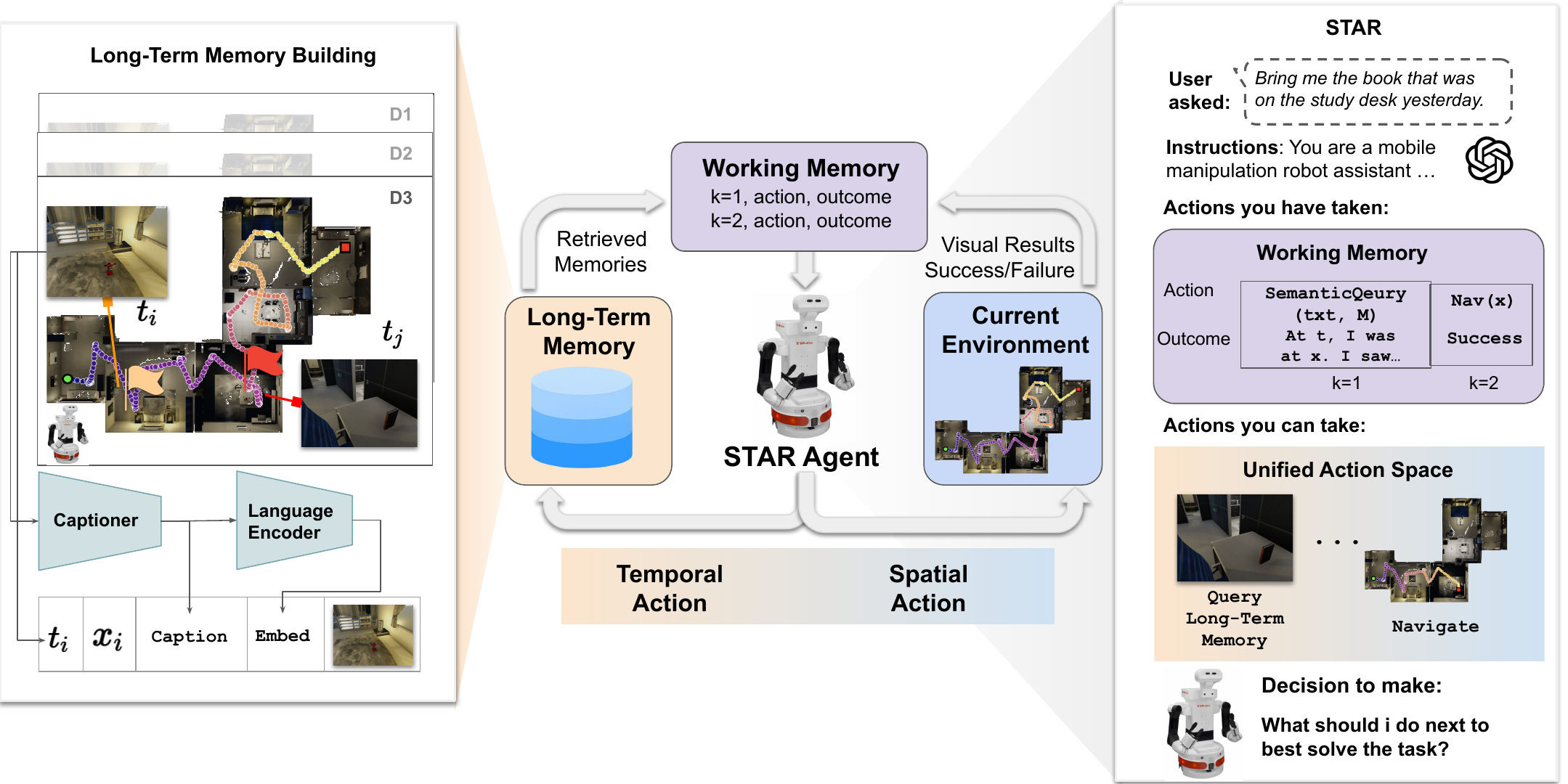}
    \caption{STAR system. The robot patrols dynamic environments over multiple days to build a non-parametric long-term memory of past observations (left). When the user requests “bring me the book that was on the study desk yesterday”, the agent initializes its working memory with the task. Guided by this working memory, STAR chooses actions from a unified space: recalling past observations (search in time) or probing the current environment through navigation, perception, and manipulation (search in space). Each outcome updates the working memory, and the loop continues until the robot successfully retrieves the target object.}
    \label{fig:sf}
    \vspace{-2em}
\end{figure*}

\subsection{System Overview}

To solve the open-world retrieval problem, robot needs to search in time by recalling historical evidence in $M$ to form hypotheses about the object's current states, and search in space by acting in the world to validate those hypotheses and acquire observations that guide where to search next.

To realize this capability, we present \methodname{} shown in Fig.~\ref{fig:sf}.
It uses a language model as the policy, which allows the robot to handle open-vocabulary object descriptions and understand spatial and temporal cues.
To search over $M$, a naive approach would condition the entire $M$ in the model's context, but this is impractical:
$M$ spans long histories that can exceed the model’s context window, and instruction-irrelevant details can distract the policy.
Accordingly, \methodname{} is organized around two complementary memories: (i) an efficiently searchable long-term memory $M$ that supports fast retrieval of task-relevant evidence,
and (ii) a working memory $H$ that only retains task-relevant information, initialized at task time $T$ and updated online with evidence retrieved from $M$ and observation streams $S_{T: T+k}$.
As a result, the LLM policy needs only to act based on the information in the working memory and the instruction, allowing it to focus on task-relevant information:
\begin{align}
\label{eqn:approx-problem}
a_k \sim \pi(a |\ell, H_k) \approxeq \pi(a |l, M, S_{T:T+k}),
\end{align}
where we use the timestep $k$ to denote the policy execution step.
To decide when to search in time (query $M$) versus when to search in space (take embodied actions), we expose a unified action space to the LLM policy.
At each step, the policy selects either a temporal action (retrieve from $M$) or a spatial action (navigate, observe, or manipulate). The resulting evidence is used to update the working memory.

In the following, we provide details on: 
(1) the design of the efficiently searchable long-term memory $M$ in Section~\ref{sec:long-term-memory},
(2) the process of updating the working memory $H$ with temporal and spatial actions in Section~\ref{sec:working-memory},
(3) and how the policy $\pi(a\mid \ell, H)$ uses the updated working memory to make decisions in Section~\ref{sec:working-mem-execution}.

\subsection{Non-Parametric Long-term Memory}
\label{sec:long-term-memory}

We now detail the design of the long-term memory.
This memory stores past observations in a non-parametric form in a vector database, together with lightweight descriptors that make them searchable during task execution.
Following prior works~\cite{anwar2025remembr,xie2025embodiedraggeneralnonparametricembodied}, we maintain multiple vector indices that allow the agent to query memory along different modalities:
\begin{itemize}
    \item \textit{Semantic index:} Embeddings $\texttt{embed}(o_t)$ of observations are obtained by captioning each observation and embedding the caption text. This allows for semantic queries by text.
    \item \textit{Temporal index:} Timestamp vectors $t$, which support queries about what was observed around a given time.
    \item \textit{Spatial index:} Robot poses $x_t$, which allow robots to find out visible objects around that location.
\end{itemize}
This structure makes the memory naturally extensible: new modalities can be supported by adding indices derived from their vector representations, without altering the underlying storage structure.

Building on prior approaches, a key design choice in our memory is to also retain the raw visual observations~$o_t$ alongside their indices. If the agent determines that the embedding $\texttt{embed}(o_t)$ misses task-relevant details, it can revisit the original observation $o_t$ to recover the necessary information.
Each record at timestep $t$ stores its temporal, spatial, semantic, and raw components in tuples: $m_t = (t, x_t, \texttt{embed}(o_t), o_t)$.
In the next section, we introduce how these memory records from $M$ are selectively retrieved by the temporal action to update the working memory $H$.

\subsection{Working Memory Update with Spatio-temporal Actions}
\label{sec:working-memory}

The working memory $H$ contains the \emph{task-relevant information} that policy uses to make decisions.
It is initialized at task time $T$ and updated by (i) the evidence selectively retrieved from the long-term memory~$M$,
and (ii) observations after taking a spatial action since the task time $T$. 
Formally, the working memory at timestep $k$ is represented as an action-outcome trajectory
$H_k = (a_{1:k-1}, y_{1:k-1})$,
where $a_k$ denotes the action taken at $k$, and $y_k$ indicates the execution outcome of $a_k$.
Each action is a tool-argument pair, where the tools are implemented as callable programs.
Let \(\mathcal{F}=\{f_1,\ldots,f_N\}\) denote the available tools exposed to the LLM policy and $(\Theta=\{\Theta_1, \ldots, \Theta_N\}$ the tool-specific parameter spaces.
An action at time $k$ is \(a_k=(f,\theta)_k\) with \(f_i\in\mathcal{F}\) and \(\theta_i\in\Theta_i\).
We call $\mathcal{A} \coloneqq (\mathcal{F}, \Theta)$ the unified action space because the $\mathcal{F}$ contains both temporal and spatial search tools.

To update the working memory with task-relevant information, the LLM policy examines the current $H_k$ and selects the tool $f$ and its argument $\theta$. 
The action is then executed either in the long-term memory $M$
or in the environment to retrieve information to be appended to $H_k$. 
In this work, we use three types of tools (temporal, spatial, or semantic) for the temporal actions.
Their function arguments $\theta$ are represented as text output by the LLM policy.
Internally, these functions convert $\theta$ into the corresponding vector representation. 
These vector representations are then used to compute the top matches of the memory records in $M$ ranked by standard vector similarity metrics:
\begin{align}
\label{eq:mem-query}
y_k = \{m_i : i \in I_r(\theta; M) \,\} \coloneq f(\theta; M),
\end{align}
where \(f\) is the selected temporal action and \(I_{r}(\theta; M)=\{i_1,\ldots,i_r\}\) are the indices of the top-\(r\) records in \(M\) nearest to the vector representation derived from the function argument \(\theta\) under the chosen similarity metric.
An example of the temporal tool $\texttt{SemanticQuery}$ is shown in Fig.~\ref{fig:sf}. 
It converts its function argument $\theta$ (caption to the image observation) to the language embedding using the LLM policy.
The embedding is then compared against the semantic index $\texttt{embed}(o_t)$ in the long-term memory $M$ to find the top-most matches. 
These records ground the temporal references, 
such as “the cup you saw yesterday” 
, enabling the agent to identify where and how an object was last observed.
 
Spatial actions allow the agent to \emph{look and act} in the current environment. 
Tools $f$ for the spatial actions are the robot skills (e.g., navigation, grasping, invoking a detection module), and the arguments $\theta$ are the skill parameters  (target pose, grasping point).
After the policy executes the spatial action, it receives perception feedback from the environment.
Depending on the skills executed, the perception feedback can be raw sensor observations, structured detections (e.g., bounding boxes/masks with scores), or a task-level execution signal (e.g., success/failure). 
We write the overall outcome as $y_k$.
Effectively, the agent queries the real world for evidence: the outcome $y_k$ verifies whether the current observation matches memory, refreshes knowledge with up-to-date observations, and tests action feasibility in a dynamic environment.

With the retrieved temporal and spatial outcomes, we update the working memory by appending these outcomes to it, along with the executed action, $a_k$:
\begin{align}
\label{eq:working-mem-update}
H_{k+1} = H_k \oplus (a_k, y_k) = (a_{1:k}, y_{1:k})
\end{align}

\subsection{Action Selection with Working Memory}
\label{sec:working-mem-execution}

The goal of the LLM policy is to select the tool $f$ and its argument $\theta$ from the unified action space to gather information and retrieve an object instructed in $\ell$ within a fixed budget $K$.
To make the policy aware of the budget, we also provide the policy a \emph{remaining budget} $R_k \;=\; (T{+}K) - k$ explicitly in its context so that planned actions respect the constraint (prefer less exploratory actions when $R_k$ is small). 
By conditioning on the current working memory and the remaining budget, the policy $a_k \sim \pi(a \mid \ell, H_k, R_k)$ selects an action \(a_k=(f,\theta)_k\), executes it, observes the outcome \(y_k\), and updates the memory to \(H_{k+1}\) as in Eq.~\ref{eq:working-mem-update}.
The execution loop stops if the object is retrieved (completes the instruction) or when $R_{k}{=}0$ (budget exhausted).

\section{\benchmarkname{}: A Benchmark for Spatiotemporal Object Search}

\begin{figure}[!t]

\centering
\scriptsize
\includegraphics[width=1.0\linewidth]{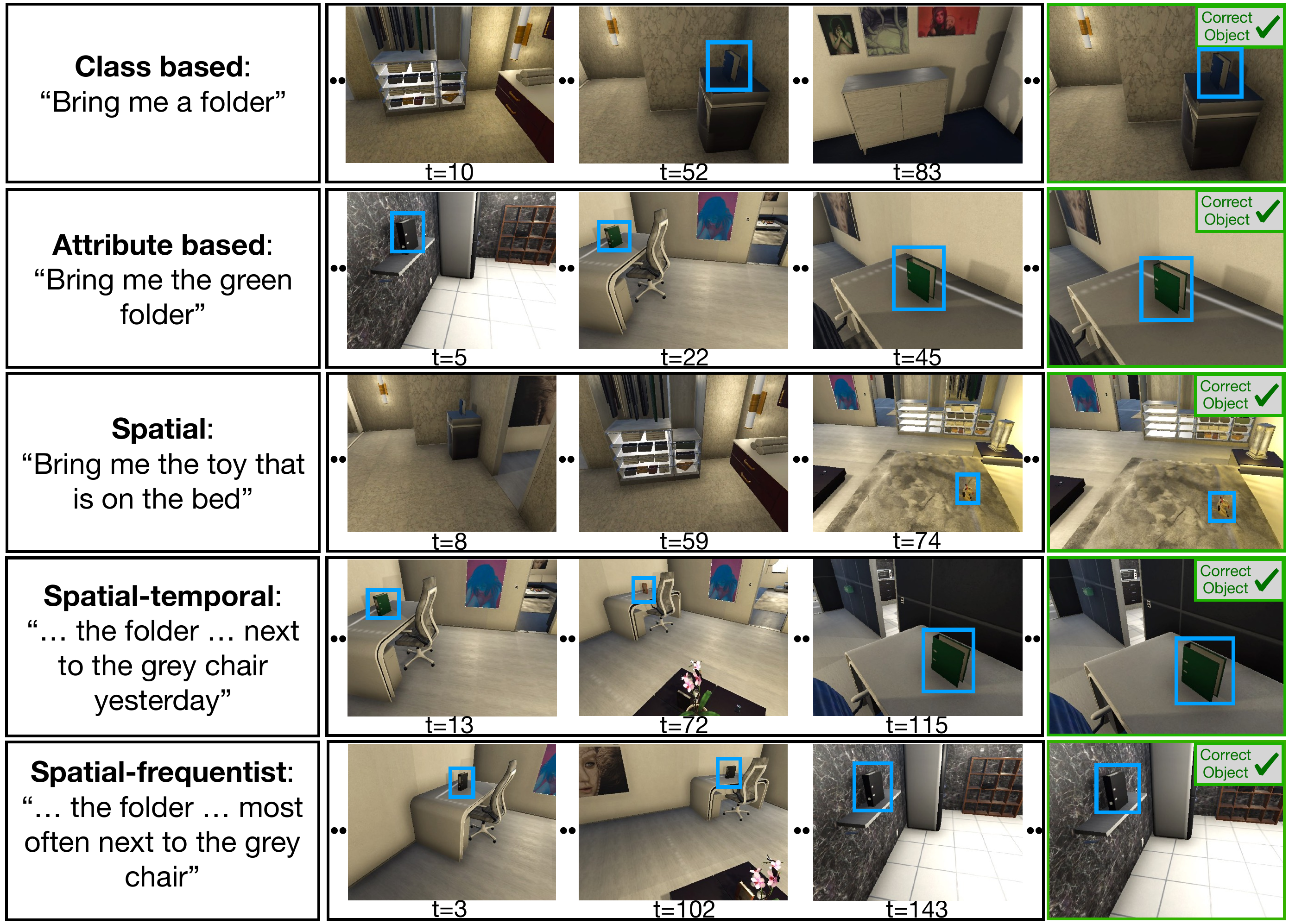}
\caption{\textbf{Task families in \benchmarkname{}.} Each row shows one task family with its instruction (left), the agent’s prior observations (middle), and the correct target object (right). Bounding boxes indicate locations of interest including target object positions. \textbf{Curr. Observation} (Current agent observation). In the \emph{class-based} case, the folder seen at $t=52$ is the target object. In the \emph{attribute-based} case, the instruction is “find the green folder”; although a black folder is visible at $t=5$, the correct target is the green folder first observed at $t=22$. In the \emph{spatial} case, the toy observed on the bed at $t=74$ is the target. In the \emph{spatial-temporal} case, the folder was seen one day earlier next to the grey chair at $t=8$ later moved, and its new position at $t=115$ is the target. In the \emph{spatial-frequentist} case, the black folder most often found next to the grey chair but last observed at $t=143$ in a new location is the target.}
\label{fig:starbench}
\end{figure}

Existing object-search benchmarks largely assume static environments; we instead evaluate open-world retrieval in homes that evolve over time. We introduce \emph{STARBench}, built on VirtualHome’s apartment-style scenes and Unity3D simulator, which provides interactive objects and articulated receptacles across multiple furnished apartments~\cite{puig2018virtualhome}. VirtualHome offers several distinct layouts, enabling controlled scene changes across days.

In \benchmarkname{}, each task is specified by a natural language instruction to retrieve a target object. Before task time, the agent patrols the environment, collecting egocentric observations, poses, along with captions of its visual observations that can later be used as memory.

To comprehensively evaluate agent ability, we introduce three task types. The first, \emph{Visible Object Search}, tests whether the agent can resolve references when the target object is directly accessible. The second, \emph{Interactive Object Search}, builds on the same object reference but requires embodied interaction, such as opening receptacles, to uncover hidden objects. The third, \emph{Commonsense Object Search}, goes beyond memory recall: here the target object may never have been observed, and the agent must rely on commonsense reasoning, for example inferring that a book is likely to be on a desk. Since both visible and interactive search require reasoning over memory, we further structure them into five task families (Fig.~\ref{fig:starbench}): \textit{(1) Class-based} (e.g., “find the book”), \textit{(2) Attribute-based} (e.g., “find the red mug”), \textit{(3) Spatial} (e.g., “find the mug on the table”), \textit{(4) Spatial Temporal} (e.g., “find the book that was on the desk yesterday”), and \textit{(5) Spatial Frequentist} (e.g., “find the mug that is usually by the sink”). Together, these task types and families span a spectrum from memory grounded reasoning, to embodied interaction, to generalization beyond observed experience. 

To prepare each task in \benchmarkname{} for execution at time $T$, the agent patrols the environment for $3$–$6$ days, collecting $1200$–$1500$ observations per day. Each observation $s_t = (t, o_t, x_t)$ records the timestep $t$, an egocentric camera observation $o_t$, and the agent pose $x_t$, along with the caption describing $o_t$. This sensor history $S_{\leq T} = \{s_i\}_{i=0}^T$ serves as input for constructing the long-term memory in \methodname, as described in Sec.~\ref{sec:long-term-memory}. We also provide scene graph representation of the environment for each task. Table \ref{tab:sim_benchmark} summarizes more statistics about the benchmark.

\begin{table}[tbh]
\scriptsize
\centering
\vspace{0.3em}
\begin{tabular}{lccc}
\toprule
Task Type & Visible & Interactive  & Commonsense \\
\midrule
No. of scenes   & $3$   & $3$  & $3$ \\
No. of Tasks    & $225$ & $90$ & $45$ \\
No. of Object Classes  & $5$   & $2$ & $5$   \\
 Task families & $5$& $5$ & $1$ \\
\bottomrule
\end{tabular}
\caption{\textbf{Summary of \benchmarkname{}.} The table shows the number of scenes, tasks, and objects for visible and interactive object search. We construct three task types with a total of $360$ tasks: \textbf{Visible} (Visible Object Search), \textbf{Interactive} (Interactive Object Search) and \textbf{Commonsense} (Commonsense Object Search)}
\label{tab:sim_benchmark}
\end{table}

\label{s:benchmark}
\section{Experimental Evaluation}

\begin{figure}[!t]
\centering
\scriptsize
\includegraphics[width=1.0\linewidth]{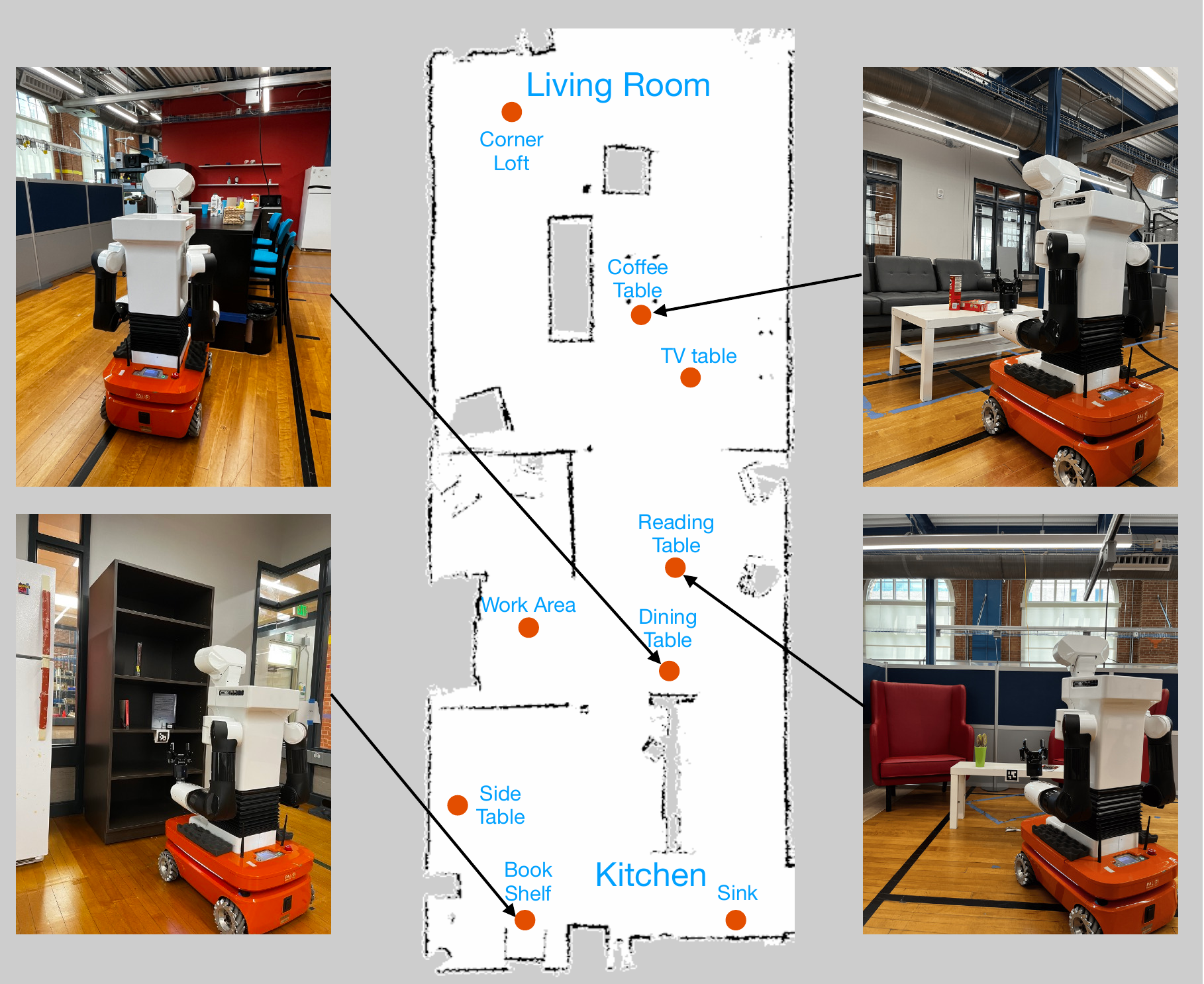}
\caption{\textbf{Mock Apartment for real-world evaluations.}{ We construct a mock apartment with a kitchen, a living room, and a study area.}}
\label{fig:real_robot_setup}
\end{figure}

\textbf{Setup:} In our experiments, we measure the performance of \methodname{} on our benchmark \benchmarkname{} and in the real world. 
We report \emph{execution success}, defined as whether the agent retrieves the user-requested object from the current environment within the step budget ($K=20$). 
We use GPT-o3~\cite{openai_o3_2025,openai_o3_systemcard_2025} as LLM backbone and query it at each step to select one action to search either in space or time.
As discussed in Sec.~\ref{sec:working-memory}, \methodname{}'s agent is equipped with tools to query past text and image observations stored in long-term memory (temporal retrieval, TR) for task-relevant information, and can also execute navigate, detect, pick, and open skills in the simulator (spatial search, S).

To separate perception errors from decision-making, we provide the agents with two modes of environmental knowledge when building their long-term memory. In the \textit{Oracle} environment mode, the agents are given ground-truth class labels of all objects, i.e., when building non-parametric memory, the captioner would have ground-truth known class labels of all visible objects, and scene graphs would store ground-truth node labels and edge relationships; for the scene-graph agent, we additionally provide ground-truth data association for nodes. 
In the \textit{Realistic} mode, no privileged information is given: agents rely entirely on predictions from state-of-the-art perception models to construct long-term memory. To isolate predicted perception errors, all agents use segmentation masks from the VirtualHome simulator for object detection during skill execution. 

We evaluate four types of solutions: 
\textbf{Random} serves as a lower bound by navigating to random locations to search for the target object. 
\textbf{SG+S} uses ground-truth scene graphs (SG) of the environment over time in its working memory, and makes a single attempt to retrieve the object (+S). 
\textbf{TR+S} uses our temporal retrieval design (TR) and then executes a one-shot plan (+S). \textbf{\methodname{}} is our full approach, combining temporal retrieval with spatial search to retrieve the object.

To evaluate the performance of \methodname{} in the real world, we deploy in on a Tiago robot in a mock household apartment environment (Fig. \ref{fig:real_robot_setup}). We followed the \benchmarkname{} procedure and created 9 tasks per type. 
A trial is considered successful if the robot leads the user to a correct landmark position where the target item can be found. 

\subsection*{Experimental Results}

With our experiments, we aim to answer a series of research questions: 

\begin{figure}[!t]
  \centering
  \includegraphics[width=\linewidth]{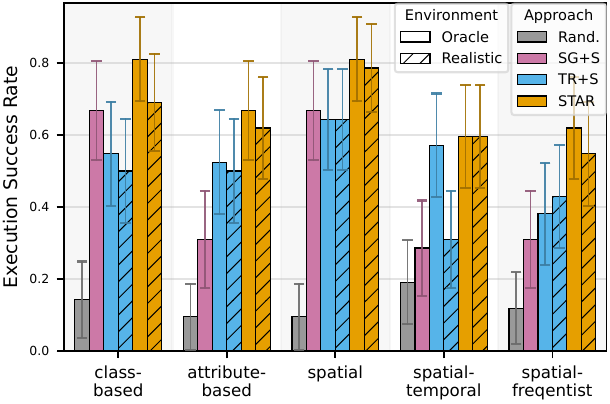}
  \caption{Execution success rates across five task types of Visible Object Search tasks in \benchmarkname{} (45 tasks per type). 
  Bars indicate approaches; hatching denotes the environment knowledge used to construct long-term memory. \textit{Oracle} builds memory with ground-truth object class labels; \textit{Realistic} builds memory from model predictions only.
  SG+S uses full scene-graph history for a one-shot attempt; TR+S queries non-parametric memory for a one-shot attempt; \methodname{} (ours) combines temporal retrieval with spatial search and achieves the highest success across task types.}
  \label{fig:results-main}
\end{figure}

\begin{figure}[!t]
\centering
\scriptsize
\includegraphics[width=1.0\linewidth]{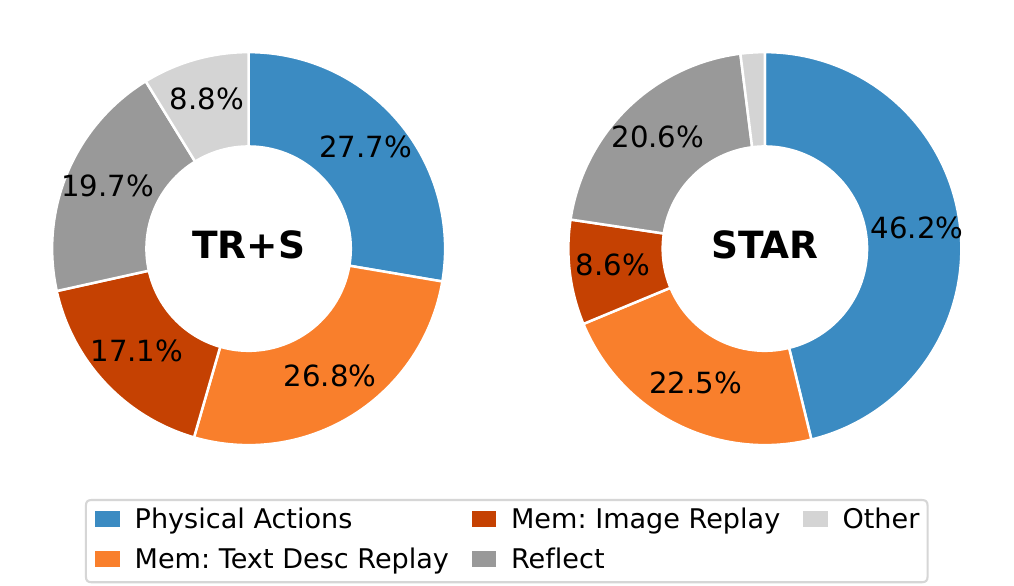}
\caption{Distribution of executed actions for all tasks.}
\label{fig:results-actions}
\end{figure}

(1) \textit{Can \methodname{} jointly reason over object attributes, space and time to identify and retrieve the correct object requested by users in dynamic environments?}
Fig~\ref{fig:results-main} reports execution success across 5 task types in visible object search. TR+S and SG+S each show complementary strengths: TR+S performs better on tasks requiring temporal recall, while SG+S retains an advantage on class-based and spatial reasoning.
\textbf{\methodname{} consistently outperforms the baselines on tasks requiring attribute, spatial, and temporal reasoning,} combining the benefits of temporal memory and spatial reasoning in a unified loop. This advantage is especially pronounced in attribute-based and spatiotemporal tasks, where reasoning about both object properties and past states is required. In addition, Fig.~\ref{fig:results-actions} shows that
\textbf{actively gathering perceptual evidence when temporal recall is insufficient helps \methodname{} retrieve objects in dynamic environments,}
indicated by \methodname{}'s larger proportion of physical actions compared to TR+S, which reduces reliance on replaying stored captions and images. By navigating and detecting when memory is uncertain or potentially outdated, \methodname{} updates stale information and acquires fresh perceptual evidence. 

\begin{table}[tbh!]
\centering
\vspace{0.5em}
\caption{Execution success rates for interactive search tasks in \benchmarkname{}. 
Task abbreviations: C = Class-based, A = Attribute-based, S = Spatial, ST = Spatial-Temporal, SF = Spatial-Frequentist.}
\begin{tabular}{lcc|cc}
\toprule
\multirow{2}{*}{\textbf{Task}} & 
\multicolumn{2}{c|}{\textbf{TR+S}} & 
\multicolumn{2}{c}{\textbf{\methodname{}}} \\
 & Oracle & Realistic & Oracle & Realistic \\
\midrule
C   & 0.56 $\pm$ 0.21 & 0.28 $\pm$ 0.19 & 0.67 $\pm$ 0.20 & 0.67 $\pm$ 0.20 \\
A   & 0.33 $\pm$ 0.20 & 0.17 $\pm$ 0.17 & 0.50 $\pm$ 0.21 & 0.61 $\pm$ 0.21 \\
S   & 0.22 $\pm$ 0.18 & 0.06 $\pm$ 0.12 & 0.50 $\pm$ 0.21 & 0.50 $\pm$ 0.21 \\
ST  & 0.17 $\pm$ 0.17 & 0.28 $\pm$ 0.19 & 0.61 $\pm$ 0.21 & 0.28 $\pm$ 0.19 \\
SF  & 0.28 $\pm$ 0.19 & 0.11 $\pm$ 0.15 & 0.61 $\pm$ 0.21 & 0.44 $\pm$ 0.21 \\
\bottomrule
\end{tabular}
\label{tab:interactive_results}
\end{table}

(2) \textit{Can \methodname{} retrieve objects that are not directly observable by interacting with the environment (e.g., opening receptacles before picking)?}
We observe that in interactive object search tasks,
\textbf{temporal recall and visual context enable \methodname{} to open the correct receptacle and retrieve hidden objects, where other agents fail.}
Since these tasks require retrieving objects stored inside similar (a common situation in household environments)
random execution fails on long action chains, and scene graphs struggle to disambiguate nearby identical receptacles.
By leveraging temporal recall of how objects became visible and the surrounding visual context, \methodname{} identifies the correct receptacle to open and successfully retrieves the target (Table~\ref{tab:interactive_results}). 

\begin{table}[t]
\centering
\vspace{0.5em}
\caption{Average number of physical actions executed per successful run. 
“Optimal” indicates the ground-truth minimal \#action required for success.}
\label{tab:results-actions}
\begin{tabular}{l l c c c c}
\toprule
\multicolumn{6}{c}{\textbf{Visible Object Search}} \\
\midrule
Method & Env. & Percep. & Nav. & Manip. & Total \\
\midrule
\multirow{2}{*}{\methodname{}} 
 & Oracle    & 1.80 & 1.62 & 1.52 & 4.93 \\
 & Realistic & 1.70 & 1.60 & 1.44 & 4.74 \\
Optimal & -- & 1 & 1 & 1 & 3 \\
\midrule
\multicolumn{6}{c}{\textbf{Interactive Object Search}} \\
\midrule
Method & Env. & Percep. & Nav. & Manip. & Total  \\
\midrule
\multirow{2}{*}{\methodname{}} 
 & Oracle    & 3.21 & 2.00 & 3.06 & 8.27 \\
 & Realistic & 3.18 & 2.11 & 2.87 & 8.16 \\
Optimal & -- & 2 & 1 & 2 & 5 \\
\bottomrule
\end{tabular}
\end{table}

(3)\textit{ How efficiently does \methodname{} make use of physical actions (navigation or interactions) to fullfil search tasks? }
Table~\ref{tab:results-actions} shows that
\textbf{\methodname{} takes about 1.5–2× the minimal (oracle-based) number of physical actions to retrieve the objects.} Different from \textit{computational actions} (reasoning, detecting objects) that can be optimized with increasing compute, physical actions are more difficult to accelerate and therefore, costly to perform; we will explore how to further optimize their use in future work.

\begin{table}[tbh]
\centering
\small
\vspace{0.3em}
\caption{Execution success rates for common sense tasks.}
\begin{tabular}{lcc}
\toprule
\textbf{Method} & \textbf{Oracle} & \textbf{Realistic} \\
\midrule
Random   & $0.10 \!\pm\! 0.13$ & -- \\
SG+S     & $0.10 \!\pm\! 0.13$ & -- \\
TR+S     & $0.19 \!\pm\! 0.16$ & $0.43 \!\pm\! 0.19$ \\
\methodname{}     & $0.57 \!\pm\! 0.19$ & $0.57 \!\pm\! 0.19$ \\
\bottomrule
\end{tabular}
\label{tab:results-common-sense}
\end{table}

(4)\textit{ Can \methodname{} leverage common sense knowledge to locate objects never observed in memory?}
When an object has never been observed, the only way to search efficiently in environments is to exploit common sense knowledge (e.g., \textit{milk is probably in the kitchen}.
As shown in Table~\ref{tab:results-common-sense}, \textbf{\methodname{} is able to find never observed objects resorting to common sense knowledge} by reasoning about likely locations. 

\begin{table}[!t]
\centering
\vspace{0.25em}
\caption{Execution success rates for real-world deployment. (9 tasks per type; 54 tasks in total.)}
\begin{tabular}{lccc}
\toprule
\textbf{Task Type} & \textbf{Random} & \textbf{TR+S} &  \textbf{\methodname{}}  \\
\midrule
Class-based         & 0.00 & 0.44 & 0.67  \\
Attribute-based     & 0.00 & 0.67 & 0.78  \\
Spatial             & 0.11 & 1.00 & 1.00  \\
Spatial-Temporal    & 0.00 & 0.56 & 0.56 \\
Spatial-Frequentist & 0.00 & 0.44 & 0.56  \\
Common-Sense        & 0.00 & 0.33 & 0.33 \\
\bottomrule
\end{tabular}
\label{tab:overall_success_real}
\end{table}

(5) \textit{Does the performance of \methodname{} in \benchmarkname{} extend to object search in the real world? }
In our evaluation in the real world, we observe that
\textbf{\methodname{} repeatedly and robustly locate objects in the environment.}
Table~\ref{tab:overall_success_real} shows that \methodname{} surpasses TR on several task types and performs well across all, while random fails almost entirely. 
\section{Conclusions}

We introduced \methodname{} and \benchmarkname{}, addressing the challenge of spatiotemporal object retrieval in dynamic environments. Our framework unifies memory queries and embodied actions, enabling agents to reason jointly over past and present states. To evaluate this setting, we presented STARBench, a benchmark that tests object search across visible, interactive, and commonsense tasks in evolving homes. Upon acceptance, we will publicly release STARBench to support further research on open-world object retrieval in dynamic environments. Looking forward, we see two promising directions. One is \emph{forgetting}: developing strategies to prune or compress memory in response to user actions while preserving what matters most. Another is \emph{policy abstraction}, where high-level primitives are distilled from low-level action traces, in the spirit of code-as-policy~\cite{liang2022code}, to better adapt to new user instructions. 








\FloatBarrier
\IEEEtriggercmd{\balance} 
\bibliographystyle{IEEEtran}  
\bibliography{references}


\end{document}